\documentclass[11pt,letterpaper]{article}
\usepackage{emnlp2016}
\usepackage{times}
\usepackage{latexsym}

\usepackage{hyperref} 
\usepackage{breakurl} 
\usepackage{basecommon}

\usepackage{balance}

\emnlpfinalcopy

\newcommand{\checkmk}{\textbullet}  %\checkmark
\newcommand{\nocheck}{}

\title{Word Ordering Without Syntax }

\author{Allen Schmaltz \and Alexander M. Rush \and Stuart M. Shieber \\
Harvard University \\
{\tt \footnotesize \{schmaltz@fas,srush@seas,shieber@seas\}.harvard.edu}}

\date{}
\newenvironment{myquote}{\list{}{\leftmargin=0.2in\rightmargin=0.2in}\item[]}{\endlist}

\begin{document}

\maketitle

\begin{abstract}
  Recent work on word ordering has argued that syntactic structure is
  important, or even required, for effectively recovering the order
  of a sentence.
  We find that, in fact, an n-gram language model with a simple
  heuristic gives strong results on this task. Furthermore, we
  show that a long short-term memory (LSTM) language model is
  even more effective at recovering order, with our basic model
  outperforming a state-of-the-art syntactic model by 11.5
  BLEU points. Additional data and larger beams yield further gains,
  at the expense of training and search time.
\end{abstract}

\section{Introduction}

We address the task of recovering the
original word order of a shuffled sentence, referred to as bag generation \cite{BrownEtAl1990}, shake-and-bake generation \cite{Brew:1992:LCO:992133.992165}, or more recently, linearization, as standardized
in a recent line of research 
as a method useful for isolating the performance of 
text-to-text generation models \cite{ZhangAndClark2011_ccg,LiuEtAl2015_transition_base_linearization,LiuAndZhang2015_empirical_comparison,ZhangAndClark2015_syntax}. The predominant
argument of the more recent works is that jointly recovering explicit syntactic
structure is crucial for determining the correct word order of the
original sentence. As such, these methods either generate or rely on
given parse structure to reproduce the order.

Independently, \newcite{Elman1990_structure_in_time} explored
linearization in his seminal work on recurrent neural networks.  Elman
judged the capacity of early recurrent neural networks via, in
part, the network's ability to predict word order in simple
sentences. He notes,

\begin{small}
\begin{myquote}
  The order of words in sentences reflects a number of
  constraints\ldots Syntactic structure, selective restrictions,
  subcategorization, and discourse considerations are among the many
  factors which join together to fix the order in which words
  occur\ldots [T]here is an abstract structure which underlies the surface
  strings and it is this structure which provides a more insightful
  basis for understanding the constraints on word order\ldots. It is,
  therefore, an interesting question to ask whether a network can
  learn any aspects of that underlying abstract structure
  \cite{Elman1990_structure_in_time}.
\end{myquote}
\end{small}

Recently, recurrent neural networks have reemerged as a powerful
tool for learning the latent structure of language. In particular, work
on long short-term memory (LSTM) networks for language modeling has
provided improvements in perplexity.

We revisit Elman's question by applying LSTMs to the
 word-ordering task, without any explicit syntactic modeling. We find that
language models are in general effective for linearization relative to existing syntactic approaches, with LSTMs in particular outperforming the
state-of-the-art by 11.5 BLEU points, with further gains observed when 
training with additional text and decoding with larger beams.

\section{Background: Linearization}

The task of linearization is to recover the original order of a
shuffled sentence.  We assume a vocabulary $\mcV$ and are given a
sequence of out-of-order phrases $x_1, \ldots ,x_N$, with $x_n \in
\mcV^+$ for $1 \leq n \leq N$. Define $M$ as the total number of
tokens (i.e., the sum of the lengths of the phrases). We consider two
varieties of the task: (1) \textsc{Words}, where each $x_n$ consists
of a single word and $M=N$, and (2) \textsc{Words+BNPs}, where base
noun phrases (noun phrases not containing inner noun phrases) are also
provided and $M \geq N$. The second has become a standard formulation in recent literature.

Given input $x$, we define the output set $\mcY$ to be all possible
permutations over the $N$ elements of $x$, where $\hat{y} \in \mcY$ is the
permutation generating the true order. We aim to find $\hat{y}$, or a
permutation close to it. We produce a linearization by
(approximately) optimizing a learned scoring function $f$ over the set
of permutations, $y^* = \argmax_{y\in\mcY} f(x, y)$.

\section{Related Work: Syntactic Linearization}

Recent approaches to linearization have been based on reconstructing the
syntactic structure to produce the word order.  Let $\mcZ$ represent
all projective dependency parse trees over $M$ words. The objective is
to find $y^*,z^* = \argmax_{y\in\mcY, z \in \mcZ} f(x, y, z)$ where
$f$ is now over both the syntactic structure and the
linearization. The current state of the art on the Penn Treebank (PTB)
\cite{DBLP:journals/coling/MarcusSM94}, without external data, of \newcite{LiuEtAl2015_transition_base_linearization} uses a
transition-based parser with beam search to construct a sentence and a
parse tree. The scoring function is a linear model $f(x,y) =
\theta^\top \Phi(x, y, z)$ and is trained with an early update
structured perceptron to match both a given order and syntactic
tree. The feature function $\Phi$ includes features on the syntactic
tree. This work improves upon past work which used best-first search
over a similar objective \cite{ZhangAndClark2011_ccg}.

In follow-up work, \newcite{LiuAndZhang2015_empirical_comparison}
argue that syntactic models yield improvements over pure
surface n-gram models for the \textsc{Words+BNPs} case. This result holds particularly on longer sentences and even when the syntactic trees used in training are of low quality. 
The n-gram decoder of this work utilizes a single beam, discarding the probabilities of internal, non-boundary words in the BNPs when comparing hypotheses. We revisit this comparison between syntactic models and surface-level models, utilizing a surface-level decoder with heuristic future costs and an alternative approach for scoring partial hypotheses for the \textsc{Words+BNPs} case.

Additional previous work has also explored n-gram models for the word ordering task. The work of \newcite{deGispertEtAl_2014_phrasebased} demonstrates improvements over the earlier syntactic model of \newcite{ZhangEtAl_2012_ccg_word_ordering} by applying an n-gram language model over the space of word permutations restricted to concatenations of phrases seen in a large corpus. \newcite{HorvatAndByrne_2014_graph_based} models the search for the highest probability permutation of words under an n-gram model as a Travelling Salesman Problem; however, direct comparisons to existing works are not provided.

\section{LM-Based Linearization}

In contrast to the recent syntax-based approaches, we use an LM directly for word
ordering. We consider two types of language models: an n-gram model
and a long short-term memory network
\cite{Hochreiter1997_LSTM}. For the purpose of this work, we define a
common abstraction for both models. Let $\boldh \in \mcH$ be the
current state of the model, with $\boldh_0$ as the initial state. Upon
seeing a word $w_i \in \mcV$, the LM advances to a new state $\boldh_i =
\delta(w_i, \boldh_{i-1})$. At any time, the LM can be queried to produce an
estimate of the probability of the next word $q(w_i, \boldh_{i-1}) \approx
p(w_i \mid w_1, \ldots, w_{i-1})$.  For n-gram language models, $\mcH$,
$\delta$, and $q$ can naturally be defined respectively as the state
space, transition model, and edge costs of a finite-state machine.

LSTMs are a type of recurrent neural network (RNN) that are conducive
to learning long-distance dependencies through the use of an internal
memory cell. Existing work with LSTMs has generated state-of-the-art
results in language modeling \cite{Zaremba14_rnn_regularization},
along with a variety of other NLP tasks.  

\algrenewcommand\algorithmicindent{1.0em}%
\begin{algorithm}[t!]
  \small
  \begin{algorithmic}[1]
    \Procedure{Order}{$x_1\ldots x_N$, $K$, $g$}
    \State{$B_0 \gets \langle (\langle \rangle, \{1, \ldots, N\}, 0, \boldh_0)  \rangle$}
    \For{$m = 0, \ldots, M-1$ }
    \For{$k = 1, \ldots, |B_m|$}
    \State{$(y, \mcR, s, \boldh) \gets B_m^{(k)}$}
    \For{$i \in \mcR$}
    \State{$(s', \boldh') \gets (s, \boldh)$}
    \For{word $w$ in phrase $x_i$}
    \State{$s' \gets s' + \log q(w, \boldh') $ } 
    \State{$\boldh' \gets \delta(w, \boldh')$}
    \EndFor{}
    \State{$j \gets m + |x_i|$ }
    \State{$B_{j} \gets B_{j} + (y + x_i, \mcR - i, s',   \boldh')$}
    \State{keep top-$K$ of $B_{j}$ by $f(x, y) + g(\mcR)$}
    \EndFor{}
    \EndFor{}
    \EndFor{}
    \State{\Return{$B_{M}$}}
    \EndProcedure{}
  \end{algorithmic}
  \caption{\label{alg:greedy} LM beam-search word ordering}
\end{algorithm}
 
In our notation we define $\mcH$ as the hidden states and cell states of a
multi-layer LSTM, $\delta$ as the LSTM update function, and $q$ as a
final affine transformation and softmax given as
$q(*, \boldh_{i-1}; \theta_q) = \mathrm{softmax}(\boldW \boldh^{(L)}_{i-1}+\boldb) $
\noindent where $\boldh^{(L)}_{i-1}$ is the top hidden layer and $\theta_q =
(\boldW, \boldb)$ are parameters. We direct readers to the work of
\newcite{Graves13_generating_sequences} for a full description of the
LSTM update. 

For both models, we simply define the scoring function as  
\[ f(x, y) =
\sum_{n=1}^N \log p(x_{y(n)} \mid x_{y(1)},\ldots, x_{y(n-1)}) \]
where the phrase probabilities are calculated word-by-word by our
language model.

Searching over all permutations $\mcY$ is intractable, so we instead
follow past work on linearization
\cite{LiuEtAl2015_transition_base_linearization} and LSTM generation
\cite{sutskever2014sequence} in adapting beam search for our generation step.  Our work differs from the beam search approach for the \textsc{Words+BNPs} case of previous work in that we maintain multiple beams, as in stack decoding for phrase-based machine translation \cite{Koehn2010-SMTbook}, allowing us to incorporate the probabilities of internal, non-boundary words in the BNPs. Additionally, for both \textsc{Words} and \textsc{Words+BNPs},
we also include an estimate of future cost in order to improve search accuracy.

Beam search maintains $M+1$ beams, $B_0, \ldots, B_M$, each containing
at most the top-$K$ partial hypotheses of that length. A partial
hypothesis is a 4-tuple $(y, \mcR, s, \boldh)$, where $y$ is a partial
ordering, $\mcR$ is the set of remaining indices to be ordered, $s$ is
the score of the partial linearization $f(x, y)$, and $\boldh$ is the
current LM state. Each step consists of expanding all next possible
phrases and adding the next hypothesis to a later beam. The full beam
search is given in Algorithm~\ref{alg:greedy}.

As part of the beam search scoring function we also include a future
cost $g$, an estimate of the score contribution of the remaining
elements in $\mcR$.  Together,
$f(x, y) + g(\mcR)$ gives a noisy estimate of the total score, which is
used to determine the $K$ best elements in the beam. In our
experiments we use a very simple unigram future cost estimate, $g(\mcR)
= \sum_{i \in \mcR} \sum_{w \in x_i}\log p(w)$.

\section{Experiments}

\begin{table}
\centering
\small
\hspace*{-2mm}\begin{tabular}{lcc}
\toprule
Model & \textsc{Words} & \textsc{Words+BNPs} \\
\midrule
\textsc{ZGen-64}\nocite{LiuEtAl2015_transition_base_linearization} & 30.9 & 49.4 \\
\textsc{ZGen-64+pos} \nocite{LiuEtAl2015_transition_base_linearization} & -- & 50.8 \\
\midrule
\textsc{NGram-64 (no $g$)} & 32.0 & 51.3  \\
\textsc{NGram-64} & 37.0 & 54.3  \\
\textsc{NGram-512} & 38.6 & 55.6 \\
\textsc{LSTM-64} &  40.5 & 60.9 \\ 
\textsc{LSTM-512} &  42.7 & 63.2 \\ 
\midrule 
\textsc{ZGen-64+lm+gw+pos} \nocite{LiuAndZhang2015_empirical_comparison} & -- & 52.4 \\
\textsc{LSTM-64+gw}  & 41.1 & 63.1   \\
\textsc{LSTM-512+gw} & 44.5 & 65.8  \\
\bottomrule
\end{tabular}
\caption{\label{table-final-results} BLEU score comparison on the PTB test set. Results from previous works (for \textsc{ZGen}) are those provided by the respective authors, except for the \textsc{Words} task. The final number in the model identifier is the beam size, \textsc{+gw} indicates additional Gigaword data. Models marked with \textsc{+pos} are provided with a POS dictionary derived from the PTB training set. }
\end{table}

\begin{table}[t]
\centering
\footnotesize
\begin{tabular}{ccccccccc}
\toprule
\textsc{bnp} & $g$ & \textsc{gw} & $1$ & 10 & 64 & 128 & 256 & 512 \\
\midrule
&& &\multicolumn{6}{c}{\textsc{LSTM}} \\
\checkmk &\nocheck & \nocheck & 41.7 & 53.6 & 58.0 & 59.1 & 60.0 & 60.6 \\
\checkmk & \checkmk & \nocheck & 47.6 & 59.4 & 62.2 & 62.9 & 63.6 & 64.3 \\
\checkmk &\checkmk &  \checkmk & 48.4 & 60.1 & 64.2 & 64.9 & 65.6 & 66.2 \\
\nocheck &\nocheck &  \nocheck & 15.4 & 26.8 & 33.8 & 35.3 & 36.5 & 38.0 \\
\nocheck& \checkmk &  \nocheck& 25.0 & 36.8 & 40.7 & 41.7 & 42.0 & 42.5 \\
\nocheck&\checkmk & \checkmk & 23.8 & 35.5 & 40.7 & 41.7 & 42.9 & 43.7 \\
\midrule
 &&& \multicolumn{6}{c}{\textsc{NGram}} \\
 \checkmk &\nocheck & \nocheck & 40.6 & 49.7 & 52.6 & 53.2 & 54.0 & 54.7 \\
 \checkmk & \checkmk & \nocheck &  45.7 & 53.6 & 55.6 & 56.2 & 56.6 & 56.6 \\
 \nocheck &\nocheck &  \nocheck & 14.6 & 27.1 & 32.6 & 33.8 & 35.1 & 35.8 \\
 \nocheck&\checkmk & \nocheck &  27.1 &  34.6 & 37.5 & 38.1 &  38.4 & 38.7 \\
\bottomrule
\end{tabular}
\caption{\label{table-internal-results} BLEU results on the validation set
  for the \textsc{LSTM} and \textsc{NGram} model with varying beam sizes, future costs, additional data, and 
  use of base noun phrases. 
} 
\vspace{-4mm}
\end{table}

\paragraph{Setup}

Experiments are on PTB with sections 2-21 as
training, 22 as validation, and 23 as test\footnote{In practice, the results in \newcite{LiuEtAl2015_transition_base_linearization} and \newcite{LiuAndZhang2015_empirical_comparison} use section 0 instead of 22 for validation (author correspondence).}. We utilize two UNK
types, one for initial upper-case tokens and one for all other
low-frequency tokens; end sentence tokens; and start/end tokens, which are treated as words, to mark BNPs
for the \textsc{Words+BNPs} task. We also use a special symbol to replace tokens that contain at least one numeric character. We otherwise train with punctuation and the original case of each token, resulting in a vocabulary containing
around 16K types from around 1M training tokens.

For experiments marked \textsc{gw} we augment the PTB with a subset of
the Annotated Gigaword corpus \cite{Napoles2012_gigaword}.  We follow
\newcite{LiuAndZhang2015_empirical_comparison} and train on a sample
of 900k Agence France-Presse sentences combined with the full PTB training set.  The
\textsc{gw} models benefit from both additional data and a larger vocabulary
of around 25K types, which reduces unknowns in the validation and test
sets.

We compare the models of
\newcite{LiuEtAl2015_transition_base_linearization} (known as
\textsc{ZGen}), a 5-gram LM using Kneser-Ney smoothing
(\textsc{NGram})\footnote{We use the KenLM Language Model Toolkit (\url{https://kheafield.com/code/kenlm/}).}, and an \textsc{LSTM}.  We experiment on the
  \textsc{Words} and \textsc{Words+BNPs} tasks, and we also experiment
  with including future costs ($g$), the Gigaword data
  (\textsc{gw}), and varying beam size.  We retrain \textsc{ZGen} using
  publicly available code\footnote{\url{https://github.com/SUTDNLP/ZGen}} to
  replicate published results. 
  
  The \textsc{LSTM} model is similar in size and architecture to the medium LSTM setup of
  \newcite{Zaremba14_rnn_regularization}\footnote{We hypothesize that additional gains are possible via a larger model and model averaging, ceteris paribus.}. Our implementation uses the
  Torch\footnote{\url{http://torch.ch}} framework and is publicly
  available\footnote{\url{https://github.com/allenschmaltz/word\_ordering}}.

  We compare the performance of the models using the BLEU metric
  \cite{PapineniEtAl2002_bleu}. In generation if there are multiple tokens of identical UNK type, we randomly replace each with possible unused tokens in the original source before calculating BLEU. For comparison purposes, we use the BLEU script distributed with the publicly available \textsc{ZGen} code.

\paragraph{Results}

\begin{figure}[t!]
    \centering
    \includegraphics[width=1.0\columnwidth]{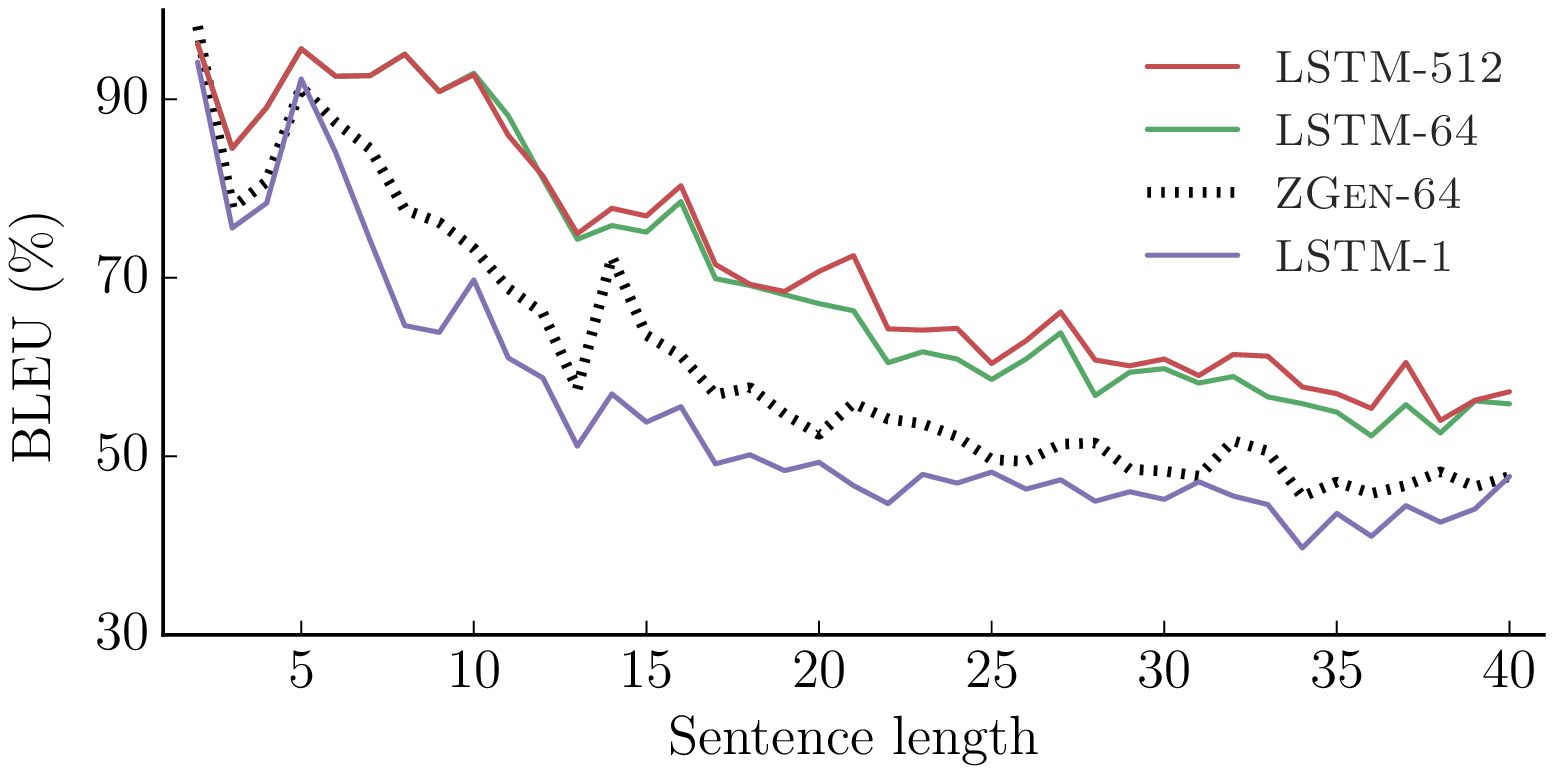}
    \includegraphics[width=1.0\columnwidth]{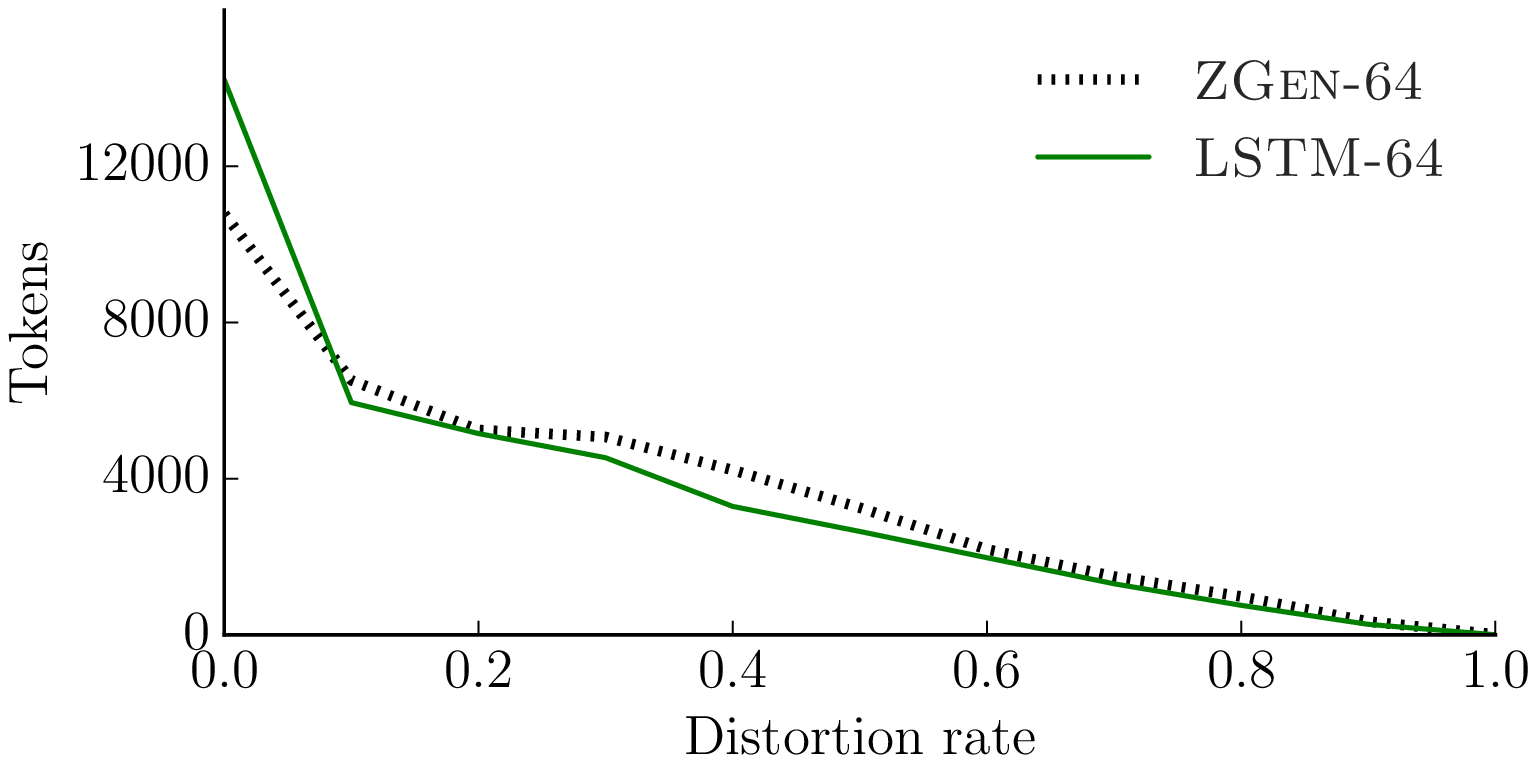}
    \caption{Experiments on the PTB validation on the \textsc{Words+BNPs} models: (a) Performance on the set by length on sentences from length 2 to length 40. (b) The distribution of token distortion, binned at 0.1 intervals.}
    \label{figure:length}
\end{figure} 
 
Our main results are shown in
Table~\ref{table-final-results}. On the
\textsc{Words+BNPs} task the \textsc{NGram-64} model scores nearly 5
points higher than the syntax-based model \textsc{ZGen-64}. The
\textsc{LSTM-64} then surpasses \textsc{NGram-64} by more than 5
BLEU points. Differences on the \textsc{Words} task are smaller, but show a similar pattern. Incorporating Gigaword
further increases the result another 2 points. Notably, the \textsc{NGram} model outperforms the combined result of
\textsc{ZGen-64+lm+gw+pos} from \newcite{LiuAndZhang2015_empirical_comparison}, which uses a 4-gram model trained on Gigaword. We believe this is because the combined \textsc{ZGen} model incorporates the n-gram
scores as discretized indicator features instead of using the
probability directly.\footnote{In work of \newcite{LiuAndZhang2015_empirical_comparison}, with the given decoder, N-grams only yielded a small further improvement over the syntactic models when discretized versions of the LM probabilities were incorporated as indicator features in the syntactic models.} A beam of 512 yields a further improvement at the cost of search time.

\begin{table}
\centering
\small
\begin{tabular}{lcc}
\toprule
Model & \textsc{Words} & \textsc{Words+BNPs} \\
\midrule
\textsc{ZGen-64}\nocite{LiuEtAl2015_transition_base_linearization}$(z^*)$ & 39.7 & 64.9  \\
\textsc{ZGen-64}\nocite{LiuEtAl2015_transition_base_linearization} & 40.8 & 65.2 \\
\midrule
\textsc{NGram-64} & 46.1 &  67.0 \\
\textsc{NGram-512} & 47.2 & 67.8  \\
\textsc{LSTM-64} &  51.3 & 71.9 \\ 
\textsc{LSTM-512} & 52.8  & 73.1 \\ 
\bottomrule
\end{tabular}
\caption{\label{table-valid-syntax} Unlabeled attachment scores (UAS) on the PTB validation set after parsing and aligning the output. For \textsc{ZGen} we also include a result using the tree $z^*$ produced directly by the system. For \textsc{Words+BNPs}, internal BNP arcs are always counted as correct. }
\end{table}

To further explore the impact of search accuracy,
Table~\ref{table-internal-results} shows the results of various models
with beam widths ranging from 1 (greedy search) to 512, and also with
and without future costs $g$.  We see that for the better models there
is a steady increase in accuracy even with large beams,
indicating that search errors are made even with relatively large
beams.

One proposed advantage of syntax in linearization models is that it
can better capture long-distance relationships. Figure~\ref{figure:length} shows results by sentence length and
distortion, which is defined as the absolute difference between a 
token's index position in $y^*$ and $\hat{y}$, normalized by $M$. The LSTM model exhibits consistently better
performance than existing syntax models across sentence lengths and
generates fewer long-range distortions than the \textsc{ZGen} model.

Finally, Table~\ref{table-valid-syntax} compares the syntactic
fluency of the output. As a lightweight test, we parse the output with the
Yara Parser \cite{RasooliEtAl15} and compare the unlabeled attachment
scores (UAS) to the trees produced by the syntactic system. We first
align the gold head to each output token. (In cases where the
alignment is not one-to-one, we randomly sample among the
possibilities.) The models with no knowledge of syntax are able to recover a higher proportion of gold
arcs.

\section{Conclusion}

Strong surface-level language models recover word order more accurately than the models trained with explicit syntactic annotations appearing in a recent series of papers. This has implications for the utility of costly syntactic annotations in generation models, for both high- and low- resource languages and domains.

\section*{Acknowledgments}

We thank Yue Zhang and Jiangming Liu for assistance in using ZGen, as well as verification of the task setup for a valid comparison. Jiangming Liu also assisted in pointing out a discrepancy in the implementation of an earlier version of our \textsc{NGram} decoder, the resolution of which improved BLEU performance.

\bibliography{gen}\balance
\bibliographystyle{emnlp2016}

\end{document}